%% file: acl_latex.tex
\documentclass[11pt]{article}

\PassOptionsToPackage{table}{xcolor}
\usepackage[preprint]{acl}

\usepackage{times}
\usepackage{latexsym}

\usepackage[T1]{fontenc}

\usepackage[utf8]{inputenc}

\usepackage{microtype}

\usepackage{inconsolata}

\usepackage{graphicx}

\usepackage{booktabs}
\usepackage{amssymb,amsmath}   
\usepackage{pifont}    
\usepackage[nameinlink,capitalise]{cleveref}
\usepackage{wrapfig}
\usepackage{subcaption}  
\usepackage{enumitem}
\usepackage{multirow}

\definecolor{darkblue}{rgb}{0, 0, 0.5}
\hypersetup{colorlinks=true, citecolor=darkblue, linkcolor=darkblue, urlcolor=darkblue}

\definecolor{pearDark}{RGB}{200, 220, 120}

\title{Speculative End-Turn Detector for Efficient Speech Chatbot Assistant}

\author{Hyunjong Ok$^{1,2}$ \quad Suho Yoo$^{2,3}$ \quad Jaeho Lee$^1$ \\
$^1$POSTECH\qquad$^2$HJ AILAB\qquad$^3$KAIST\\
\texttt{\{hyunjong.ok, uso7d0\}@gmail.com, jaeho.lee@postech.ac.kr}
}

\begin{document}
\maketitle

\input{sections/0_abstract}

\input{sections/1_introduction}

\input{sections/2_related_work}

\input{sections/3_dataset}

\input{sections/4_method}

\input{sections/5_experiments}

\input{sections/6_results}

\section{Conclusion}

We introduce the OpenETD dataset, a novel publicly available dataset for end-turn detection in spoken dialogue systems, and propose SpeculativeETD, an efficient method for real-time end-turn detection. These contributions address critical challenges in developing more natural and responsive spoken dialogue agents, particularly those powered by large language models.
The OpenETD dataset comprises synthetic and real data and is the first public dataset for end-turn detection research. This dataset enables more comprehensive training and evaluation of end-turn detection models by offering a diverse range of conversational scenarios, including artificially extended pauses and injected filler words. Also, real data further enhances the dataset's utility in assessing model generalizability in real-world conversational settings.
SpeculativeETD, represents a significant advancement in balancing efficiency and accuracy for real-time end-turn detection. Combining a lightweight GRU on-device model for on-device processing with a high-performance Wav2Vec 2.0 server-side model on the server improves accuracy while maintaining low latency. This two-stage framework effectively addresses the computational constraints of on-devices while leveraging the superior performance of larger models.

\section*{Limitations}
Our work has several limitations. First, the OpenETD dataset primarily focuses on English conversations, which may limit its applicability to other languages with different turn-taking patterns. Second, while SpeculativeETD demonstrates strong performance, the reliance on server-side computation for Gap/Pause classification introduces network latency considerations not addressed in our current evaluation. Third, our synthetic data generation, while effective, may not fully capture the complexity of spontaneous human speech patterns.

\section*{Ethics Statement}
All datasets used in this research are used exclusively for academic purposes. We release open-source processing code and, for data that cannot be directly shared due to licensing constraints, provide reproducible download scripts. We build on four data sources with distinct terms: MultiWOZ \citep{budzianowski-etal-2018-multiwoz} (Apache~2.0), Google Cloud Text-to-Speech (used under the Google Cloud Terms of Service), YouTube audio (redistributed only as URLs with download scripts; we do not rehost audio files), and the Buckeye Corpus \citep{Pitt2005TheBC} (used under its Academic License). Full licensing details are provided in \cref{app:licenses}.

\section*{Acknowledgments}
This work was supported in part by the Institute of Information \& Communications Technology Planning \& Evaluation (IITP) grant funded by the Korea government (MSIT) (Nos. RS-2024-00457882, RS-2019-II191906, 2022-0-00713) and in part by the National Research Foundation of Korea (NRF) grant funded by the Korea government (MSIT) (Nos. RS-2024-00453301, RS-2025-24873016, RS-2026-25494004).

\bibliography{custom}

\clearpage

\appendix

\input{sections/appendix}

\end{document}

%% file: sections/0_abstract.tex
\begin{abstract}
Spoken dialogue systems powered by large language models have demonstrated remarkable abilities in understanding human speech and generating appropriate spoken responses.
However, these systems struggle with end-turn detection (ETD)---the ability to distinguish between user turn completion and hesitation. This limitation often leads to premature or delayed responses, disrupting the flow of spoken conversations.
In this paper, we introduce the OpenETD Dataset, the first public dataset for end-turn detection. The OpenETD dataset consists of both synthetic speech data generated with text-to-speech models and real-world speech data collected from web sources. We also propose SpeculativeETD, a novel collaborative inference framework that balances efficiency and accuracy to improve real-time ETD in resource-constrained environments. Our approach jointly employs a lightweight GRU-based model, which rapidly detects the non-speaking units in real-time on local devices, and a high-performance Wav2vec-based model running on the server to make a more challenging classification of distinguishing turn ends from mere pauses. Experiments demonstrate that the proposed SpeculativeETD significantly improves ETD accuracy while keeping the required computations low.
\end{abstract}

%% file: sections/1_introduction.tex
\section{Introduction}

\begin{figure}[t]
    \centering
    \includegraphics[width=1\columnwidth]{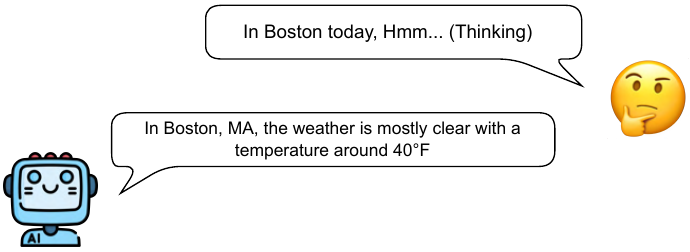}
    \caption{\textbf{End-turn detection failure example.} An example of a failure case of end-turn detection during a voice chat with GPT-4o. When the user pauses to think, the system incorrectly interprets this as the end of their turn.} \label{fig:our_task}
\end{figure}

Recent advancements in large language models (LLMs) have spurred extensive research into various LLM-based agents \citep{song2023llm, shao2024assisting, zhang-etal-2024-agent, xie2024can}. Many studies focus on integrating LLMs with spoken dialogue systems to enable more engaging human-computer interaction \citep{mitsui2023towards, yan-etal-2024-talk, ma2024language, veluri2024beyond}. However, LLM-based spoken dialogue systems face a critical challenge: when the user suddenly stops speaking, the systems often fail to discern whether the user has completed their turn or is merely pausing to think during a conversation \citep{lin2022duplex, umair2024large}. For instance, as illustrated in \Cref{fig:our_task}, when a user momentarily pauses while contemplating a specific question, the LLM may erroneously interpret this pause as the end of the turn, leading to premature and off-target responses. This task, called end-turn detection (ETD), is a challenging task that requires nuanced mechanisms based on a firm understanding of the content of the speech; na\"{i}ve cues, such as the pause duration, fails to provide meaningful information to distinguish between the turn completion and user hesitation \citep{ten2005temporal}.

Recent research has explored techniques to improve turn-taking in dialogue systems to address this issue \citep{chang2022turn, veluri2024beyond}. However, a major bottleneck is the lack of publicly available datasets. Existing datasets, such as the Fisher corpus \citep{cieri-etal-2004-fisher}, are now paid for utilizing, while studies like \citet{chang2022turn, veluri2024beyond, ma2024language} rely on private, inaccessible data.
 
To tackle this bottleneck, we construct and release the OpenETD dataset, an open-to-public dataset that is specifically designed for training and evaluating end-turn detection models. The OpenETD dataset consists of more than 120k samples with over 300 hours of conversational data. In particular, the dataset consists of both synthetic spoken conversation generated via applying text-to-speech (TTS) generation on text dialogue dataset and real-world dialogues collected from web sources.

\input{table/table1}

Beyond dataset construction, we propose an efficient end-turn detection algorithm tailored for real-time chatbot interactions under resource-constrained environments. While transformer-based audio processing models achieve strong performance, their high computational cost makes them impractical to be used in highly repetitive tasks with real-time applications, such as ETD. In contrast, models under 1M parameters can run efficiently on local devices~\citep{10.1155/2023/5870630, 10.4018/IJSWIS.330015} but suffer from degraded accuracy (see \Cref{tab:ETD_example_result}). To address this gap, we introduce SpeculativeETD, a novel collaborative inference framework that balances efficiency and accuracy. In this framework approach, we adopt both the lightweight GRU-based model \citep{chung2014empirical} and the high-performance Wav2vec-based model \citep{baevski2020wav2vec}. Here, the lightweight model operates on local devices in real time and keeps track of whether the speaker is speaking or not (either as a pause or end-of-turn). Whenever the model detects the silence, the model queries the high-performance model to make a more challenging prediction, \textit{i.e.}, telling whether the silence marks the end of the speaker's turn or not. This framework removes the need for the high-performance model to be running in real-time, dramatically saving the amount of required computations with minimal degradation in accuracy.

Our key contributions can be summarized as follows:
\begin{itemize}[leftmargin=*,topsep=0pt,parsep=1pt]
\item We release the first open-source dataset for end-turn detection, incorporating synthetic and real-world speech data.
\item We propose a novel method, SpeculativeETD, for efficient and accurate end-turn detection, leveraging a lightweight on-device model and a server-side model.
\end{itemize}

%% file: table/table1.tex
\begin{table}[t]
\centering
\resizebox{0.65\linewidth}{!}{%
\begin{tabular}{lcc}
\toprule
 Model & Param. & Acc.   \\ \midrule
 GRU & $1$M & 79.7 \\
 Wav2vec 2.0 & $94$M & 99.3 \\
\bottomrule
\end{tabular}%
}
\caption{\textbf{Model accuracy vs. size trade-off.} Experimental results on our OpenETD dataset. The 1M GRU model can achieve real-time on-device but exhibits lower accuracy.}
\label{tab:ETD_example_result}
\end{table}

%% file: sections/2_related_work.tex
\section{Related work}

\paragraph{Turn-taking.}
Turn-taking plays a crucial role in human communication, allowing speakers to transition smoothly without interruptions. Its detection is guided by linguistic, prosodic, and non-verbal cues such as pitch variation, speech rhythm, and gaze direction \citep{gravano2011turn, Levinson2015Timing}. Early turn-taking models have used finite-state machines to predict turn timing and duration \citep{raux-eskenazi-2009-finite}. 

In spoken dialogue systems, turn-taking ensures fluid human-computer interactions. Various approaches have been explored for turn boundary prediction, including end-to-end models based on automatic speech recognition (ASR) \citep{chang2022turn} and transformer-based models incorporating contextual and pragmatic cues \citep{ekstedt-skantze-2020-turngpt}. More recently, full-duplex LLM-based agents have been developed to enable simultaneous listening and responding~\citep{defossez2024moshi}. However, existing models remain computationally expensive and rely on private datasets, limiting real-time applicability and reproducibility. Our approach addresses these challenges by introducing an efficient turn-detection method and an open-source dataset.

\paragraph{Disfluency Detection.}
Disfluencies, such as interruptions, filler words, or self-corrections, occur frequently in spontaneous speech and can disrupt the natural flow of conversations. Identifying and handling these phenomena is crucial for maintaining coherent interactions in spoken dialogue systems. Traditional approaches primarily rely on text-based detection, analyzing transcriptions with span classification models to improve accuracy \citep{ghosh2022spanclassificationstructuredinformation}. Although effective, these methods depend on ASR, which introduces latency and errors. Recent studies have explored direct speech-based disfluency detection \citep{zhou2024yolostutterendtoendregionwisespeech} and integrated frameworks that combine ASR with disfluency removal for smoother real-time processing \citep{lou2020endtoendspeechrecognitiondisfluency}. However, these methods often struggle with spontaneous speech variations and real-time efficiency.

\input{figure/figtex/datapipeline}

\paragraph{Chatbot Agents.}
LLMs have advanced chatbot systems, enabling more natural spoken dialogue interactions \citep{lin2022duplex, rubenstein2023audiopalm}. Traditional chatbots rely on turn-based exchanges, whereas recent models explore real-time processing to handle overlapping speech, interruptions, and backchanneling~\citep{veluri2024beyond, ma2024language}. 

However, LLM-driven agents often suffer from high computational costs due to their autoregressive inference paradigm, leading to slow response times. Additionally, while some models aim to adaptively handle turn-taking, they frequently miss fine-grained conversational cues like hesitations and mid-turn pauses, resulting in suboptimal interactions \citep{umair2024large}. Another challenge is the reliance on proprietary datasets, restricting accessibility for further improvements \citep{cieri-etal-2004-fisher}. To address these challenges, we incorporate speculative inference for efficient real-time processing and introduce an open-source dataset for end-turn detection, fostering further research in spoken dialogue systems.

%% file: figure/figtex/datapipeline.tex
\begin{figure*}[t]
    \centering
    \includegraphics[width=1\linewidth]{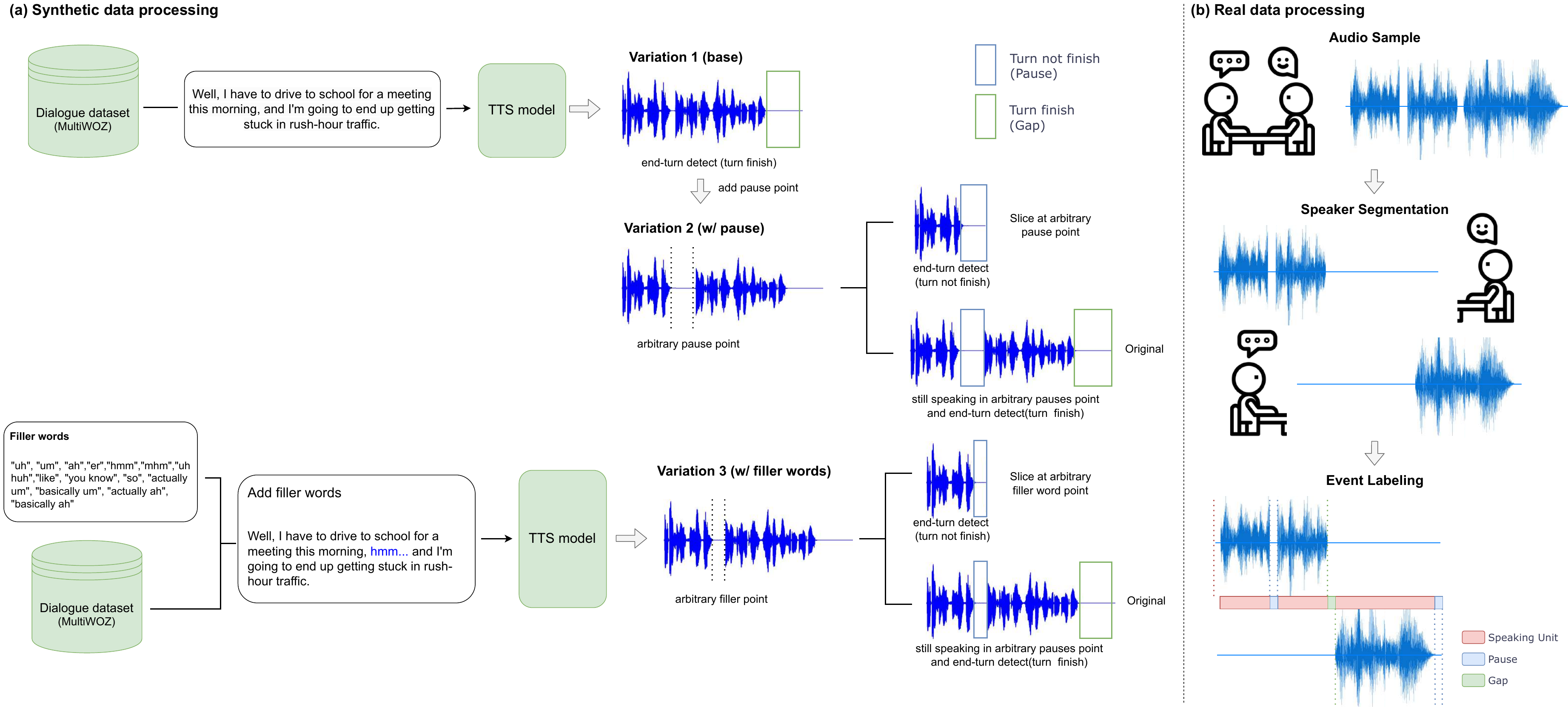}
    \caption{\textbf{Data generation methodology overview.} (a) The synthetic data pipeline converts text-based dialogue data into three types of speech variations using a text-to-speech (TTS) system. (b) The real data processing pipeline involves collecting and processing speech data from online sources.} \label{fig:DataPipeline}
\end{figure*}

%% file: sections/3_dataset.tex
\section{OpenETD: End-Turn Detection Dataset}\label{sec:dataset}

Before introducing the dataset, we first formally describe the task of end-turn detection (ETD). The goal of ETD is to infer from the given speech whether the speaker has finished speaking or not. More precisely, at each time $t$, the speaker is at one of the three states. (1) \textit{Speaking unit (SU)}: The speaker is in speech. (2) \textit{Pause}: The speaker is not in speech but is only at a brief pause and intends to keep on talking. (3) \textit{Gap}: The speaker has finished talking, marking the end of his or her turn.

The OpenETD dataset consists of multiple audio speech files labeled with the ternary segmentation information. That is, the label of each audio file is a sequence of the (state, starting time) pair of each same-state segment, \textit{e.g.,}
\begin{equation}
\begin{aligned}
\Big(&(\mathrm{SU},0\mathrm{s}), (\mathrm{Gap},12.3\mathrm{s}), (\mathrm{SU},13.7\mathrm{s}),\\
     &(\mathrm{Gap},15.0\mathrm{s}), \ldots \Big).
\end{aligned}
\end{equation}

We construct the OpenETD dataset using both synthetic (for both training and evaluation) and real (for evaluation only) conversational audio segments.

\input{table/table_synthetic_stats}
\input{table/table_real_stats}
\subsection{Synthetic data generation}\label{ssec:synthetic}

We have constructed the synthetic audio dataset of multi-turn conversation by applying the TTS (text-to-speech) model to the existing text dialogue dataset. In particular, we have used the MultiWOZ corpus \citep{budzianowski-etal-2018-multiwoz} as the base dataset. As the TTS model, we have used Google Cloud Text-to-Speech. 

An important thing to note here is that a na\"{i}ve text-to-speech translation of text dialogues will generate data consisting only of speaking units and gaps, as the textual cues for the ``pause'' are typically missing in a text dialogue. Thus, we artificially insert the pauses to the conversation in three different styles (\Cref{fig:DataPipeline}(a)):
\begin{itemize}[leftmargin=*,topsep=0pt,parsep=1pt]
\item \textbf{Base}: We do not insert any pause.
\item \textbf{w/ Pause}: We randomly select tiny hesitations---typically inserted by TTS models to mimic human breathing \citep{braunschweiler2011automatic, hwang2023pausespeech, yang2024frame}---and extend them into pauses. Here, we sample each pause duration $X$ from an Erlang distribution $\mathrm{Erlang}(k, \lambda)$ with shape $k{=}3$ and rate $\lambda{=}4.29$, truncated to $[0.1, 3.0]\,\mathrm{s}$. This yields a right-skewed distribution with mean $0.70\,\mathrm{s}$ and peak around $0.4$--$0.6\,\mathrm{s}$, matching the typical pattern of conversational pauses (many short, few long). Statistical validation against real data is provided in \cref{app:pause_stats}. For some randomly picked samples, we also remove the segments that follow the pause, generating samples without a gap at the end of the audio file.
\item \textbf{w/ Filler words}: We randomly inject the filler words, \textit{e.g.}, ``um'' or ``uh,'' at arbitrary locations in the text dialogue. After the TTS generation, we add a pause after the filler words. Again, we randomly select a fraction of samples and remove the segments after the pause to generate a sample that does not end with a gap.
\end{itemize}

\begin{figure*}[t]
    \centering
    \includegraphics[width=1\linewidth]{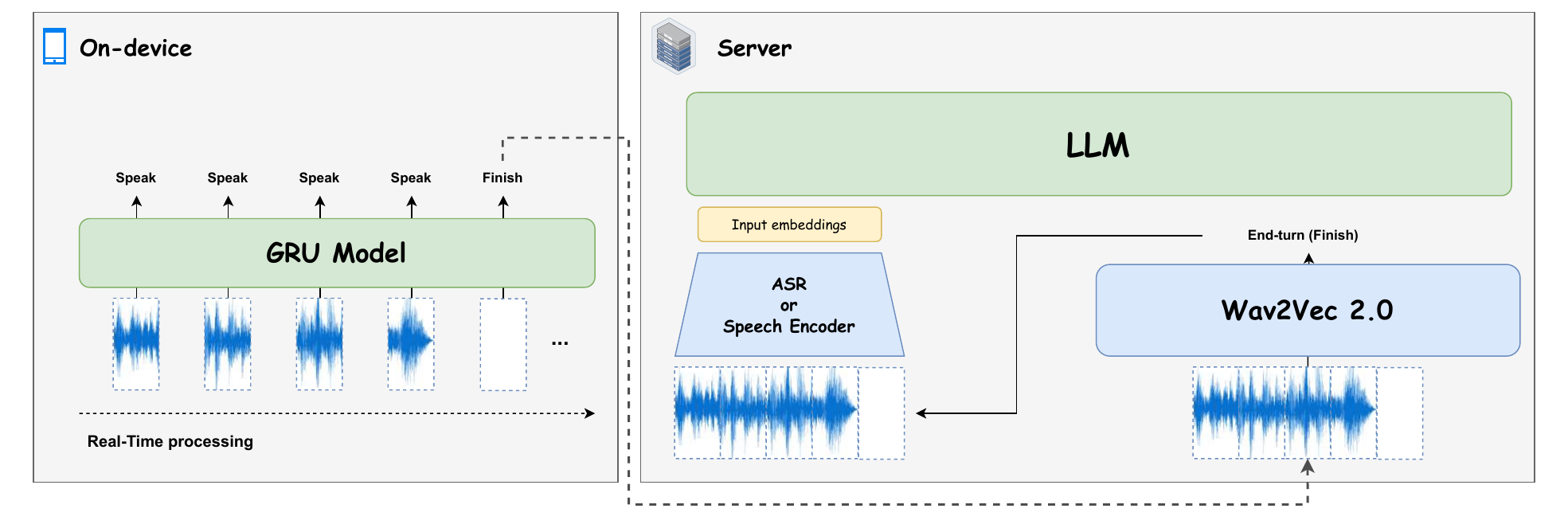}
    \caption{\textbf{SpeculativeETD framework overview.} A lightweight model (a 1M-parameter GRU in this example) operates as the on-device model, enabling real-time processing. A high-performance model (a 94M-parameter Wav2Vec 2.0 base model in this case) serves as the server-side model, verifying the predictions of the on-device model.} \label{fig:Our_Model}
\end{figure*}

\subsection{Real data generation}\label{ssec:real}

We have also collected real spoken conversation data from YouTube and Buckeye speech corpus \citep{Pitt2005TheBC}; we have selected these sources as they contain many spontaneous, natural dialogues between two speakers with frequent turn-takings. For YouTube, we used broad queries (\textit{e.g.}, \textit{``english conversation podcast''}, \textit{``podcast''}) via \texttt{yt-dlp}, retrieving up to $1{,}000$ relevance-ranked results per query, and kept only two-speaker conversations after diarization (monologues and multi-party discussions were filtered out). To ensure that the collected samples involve exactly two speakers, we have applied a pretrained speaker diarization toolkit \citep{Bredin23} to each audio file and filtered out those with more than two identified speakers.

Given conversation data, we again utilized the speaker diarization tool to generate segmentations of the speech of two different speakers (\Cref{fig:DataPipeline} (b)). Any silent interval between speech segments that exceed 200ms is labeled as either pause or gap, depending on whether the neighboring speech segments belong to the same speaker or not; here, the 200ms criterion follows the commonly used thresholds in the turn-taking literature \citep{doi:10.1073/pnas.0903616106,HELDNER2010555,nguyen2022generativespokendialoguelanguage,li2022ispeakpredictinginitiation}. Similarly to the synthetic dataset, we have randomly selected some samples and removed the audio segment that follows a randomly chosen gap or pause.

\subsection{Other details}
\textbf{Splits and statistics.} The default train/dev/test split of the synthetic data follows that of the MultiWOZ dataset \citep{budzianowski-etal-2018-multiwoz}. For the real data, we partition the collected YouTube and Buckeye conversations into disjoint train/dev/test splits at the file level, so that no source conversation appears in more than one split. The statistics of the synthetic and real datasets are given in \cref{synth_statistics} and \cref{real_statistics}, respectively.

\textbf{Human validation.} To quantify the label noise introduced by the automatic diarization + $200\,\mathrm{ms}$ rule, we conducted a human validation study on a subset of real data (three annotators, $96$ clips). Annotators reached $85.4\%$ agreement with the automatic labels and rated diarization quality at $4.17/5.0$ on average; full results are in \cref{app:human_validation}.

\textbf{Files.} For real data crawled from YouTube, we do not directly share the audio files to avoid license issues. Instead, we release the code and public URL to download the audio files.

%% file: table/table_synthetic_stats.tex
\begin{table*}[t]
\centering
\resizebox{1.5\columnwidth}{!}{%
\begin{tabular}{l r r r r}
\toprule
 & \textbf{All} & \textbf{V1 (base)} & \textbf{V2 (w/ pause)} & \textbf{V3 (w/ filler words)} \\
\midrule
\multicolumn{5}{c}{\textbf{Total}}\\
\midrule
\textbf{\# Samples}             & 122{,}481 & 28{,}368 & 47{,}076 & 47{,}037 \\
\textbf{Total duration (h)}     & 148.26    & 28.66    & 69.60    & 49.99 \\
\textbf{Average duration (s)}   & 4.36      & 3.64     & 5.32     & 3.83 \\
\midrule
\multicolumn{5}{c}{\textbf{Train}}\\
\midrule
\textbf{\# Samples}             & 96{,}773 & 22{,}321 & 37{,}246 & 37{,}206 \\
\textbf{Total duration (h)}     & 116.83   & 22.49    & 54.87    & 39.47 \\
\textbf{Average duration (s)}   & 4.35     & 3.63     & 5.30     & 3.82 \\
\midrule
\multicolumn{5}{c}{\textbf{Dev}}\\
\midrule
\textbf{\# Samples}             & 12{,}840 & 3{,}008 & 4{,}916 & 4{,}916 \\
\textbf{Total duration (h)}     & 15.75    & 3.06    & 7.37    & 5.32 \\
\textbf{Average duration (s)}   & 4.42     & 3.66    & 5.40    & 3.90 \\
\midrule
\multicolumn{5}{c}{\textbf{Test}}\\
\midrule
\textbf{\# Samples}             & 12{,}868 & 3{,}039 & 4{,}914 & 4{,}915 \\
\textbf{Total duration (h)}     & 15.68    & 3.11    & 7.36    & 5.20 \\
\textbf{Average duration (s)}   & 4.39     & 3.69    & 5.39    & 3.81 \\
\bottomrule
\end{tabular}%
}
\caption{\textbf{OpenETD synthetic data spans 148\,h across three pause variations and disjoint train/dev/test splits.} Statistics of the synthetic data grouped by pause variation: V1 is the base dialogue without explicit pauses, V2 injects pause silences, and V3 additionally inserts filler words before the pause. Splits follow the MultiWOZ partitioning of source dialogues (no speaker overlap across splits). ``V'' denotes different variations of pauses.}
\label{synth_statistics}
\end{table*}

%% file: table/table_real_stats.tex
\begin{table*}[t]
\centering
\resizebox{1.5\columnwidth}{!}{%
\begin{tabular}{l r r r r r r}
\toprule
 & \textbf{All} & \textbf{\{SU,Pause\}} & \textbf{\{SU,Gap\}} & \textbf{\{SU,Pause,Gap\}} & \textbf{Total Pause} & \textbf{Total Gap} \\
\midrule
\multicolumn{7}{c}{\textbf{Total}}\\
\midrule
\textbf{\# Samples}     & 8{,}987 & 3{,}756 & 649 & 4{,}582 & - & - \\
\textbf{Duration (h)}   & 165.75 & 43.98 & 8.08 & 113.69 & 7.78 & 3.47 \\
\midrule
\multicolumn{7}{c}{\textbf{YouTube}}\\
\midrule
\textbf{\# Samples}     & 6{,}089 & 2{,}643 & 514 & 2{,}932 & - & - \\
\textbf{Duration (h)}   & 133.56 & 35.58 & 6.96 & 90.97 & 4.82 & 2.19 \\
\midrule
\multicolumn{7}{c}{\textbf{Buckeye}}\\
\midrule
\textbf{\# Samples}     & 2{,}898 & 1{,}113 & 135 & 1{,}650 & - & - \\
\textbf{Duration (h)}   & 32.21 & 8.40 & 1.13 & 22.72 & 2.96 & 1.27 \\
\midrule
\multicolumn{7}{c}{\textbf{Train}}\\
\midrule
\textbf{\# Samples}     & 6{,}290 & 2{,}624 & 449 & 3{,}217 & - & - \\
\textbf{Duration (h)}   & 117.16 & 31.12 & 5.47 & 80.58 & 5.54 & 2.46 \\
\midrule
\multicolumn{7}{c}{\textbf{Dev}}\\
\midrule
\textbf{\# Samples}     & 899 & 363 & 67 & 469 & - & - \\
\textbf{Duration (h)}   & 16.23 & 3.60 & 0.91 & 11.72 & 0.74 & 0.34 \\
\midrule
\multicolumn{7}{c}{\textbf{Test}}\\
\midrule
\textbf{\# Samples}     & 1{,}798 & 769 & 133 & 896 & - & - \\
\textbf{Duration (h)}   & 32.36 & 9.26 & 1.70 & 21.40 & 1.50 & 0.67 \\
\bottomrule
\end{tabular}%
}
\caption{\textbf{OpenETD real data covers 166\,h of YouTube and Buckeye conversations with disjoint train/dev/test splits.} Statistics of the real data stratified by source platform and segment composition (whether an audio file contains speaking units with Pause, Gap, or both). ``Total Pause'' and ``Total Gap'' report the aggregate duration of the respective silence intervals obtained from the diarized timestamps.}
\label{real_statistics}
\end{table*}

%% file: sections/4_method.tex
\section{Method}\label{sec:method}

In this section, we describe \textit{SpeculativeETD}, a framework that achieves a favorable tradeoff of computational efficiency and accuracy for end-turn detection.

For a smooth real-time conversation, one needs to run the end-turn detector repetitively at a rapid frequency. Thus, it is desirable to deploy an ETD model with an extremely small computational footprint. However, lightweight models---\textit{e.g.}, a recurrent model with a single GRU layer---tend to suffer from a low detection accuracy. On the other hand, heavier transformer-based models typically require a large computational cost despite being highly accurate in the end-turn detection (see \Cref{sec:results}).

To enjoy the best of both worlds, the SpeculativeETD is designed as a collaborative inference framework that utilizes both a lightweight model (say, GRU), which runs on-device, and a heavier model (say, Wav2vec 2.0), which runs on the server. In a nutshell, the proposed SpeculativeETD operates in two stages (\Cref{fig:Our_Model}):
\begin{enumerate}[leftmargin=*,topsep=0pt,parsep=1pt]
    \item \textbf{On-device}: The lightweight recurrent model processes the streaming audio signal frame-by-frame (or chunk-by-chunk) and detects whether the user's speech has finished. In other words, the model conducts a binary classification between the Speaking Unit (SU) and non-SU (Gap and Pause). As this task is relatively easier than distinguishing between gaps and pauses, we can achieve high accuracy with tiny computational costs and latency.
    \item \textbf{Server-side}: Given a speech segment that ended with a silence, the heavier transformer-based model (e.g., a Wav2Vec 2.0) predicts whether the silence indicates the gap or the pause. Whenever the gap is detected, the language model will be called for generating an appropriate response to the speaker.
\end{enumerate}

The name of the method---SpeculativeETD---comes from its structural resemblance to the \textit{speculative decoding} \citep{leviathan2023fast}, a popular decoding algorithm for large language models that use a smaller and faster drafter model for autoregressive generation and larger models for parallel verification. We note, however, that our method critically differs from the speculative decoding in the sense that the small model and large model do not predict the same output classes in our framework; rather than playing the role of a verifier, the large model is used for making a more difficult fine-grained prediction, conditioned on the predictions of the small model.

\paragraph{Advantages.} The key advantages of this framework are twofold: (1) The trickier decision that requires a large computation (\textit{i.e.}, telling ``Gap'' vs. ``Pause'') can happen only once per each consecutive segment of silence, not at every time frame. As we will see in \Cref{sec:results}, this helps save the computation more than 10$\times$ with only a small degradation prediction accuracy. (2) The two-stage pipeline allows us to utilize on-device computation by placing the lightweight model on the edge. By doing so, we no longer need the communication between the edge device and the server to be done in a continuous manner; instead, we can communicate once every silence.

\paragraph{Inference Protocol.}
During real-time operation, the on-device model processes each new $100\,\mathrm{ms}$ chunk in constant time. The server-side Wav2vec 2.0 is triggered once the on-device GRU predicts non-SU (silence) for at least $200\,\mathrm{ms}$, \textit{i.e.}, two consecutive $100\,\mathrm{ms}$ chunks. Only the accumulated audio from the start of that silence segment is sent to the server. The server model then classifies the segment as Gap or Pause, determining whether an end-turn event has occurred or if streaming should continue. The $200\,\mathrm{ms}$ trigger follows the standard turn-taking threshold \citep{doi:10.1073/pnas.0903616106, HELDNER2010555}.

%% file: sections/5_experiments.tex
\section{Experimental Setup}

We now empirically validate the usefulness of the constructed OpenETD dataset (\cref{sec:dataset}) and the effectiveness of the proposed method, SpeculativeETD (\cref{sec:method}).

\textbf{Task.} We consider two tasks for the OpenETD:
\begin{itemize}[leftmargin=*,topsep=0pt,parsep=1pt]
\item \textit{Binary classification}: When a speech segment comes as an input, the detector reads the whole audio and determines whether the speech segment has ended (Gap) or not (Pause).
\item \textit{Real-time audio segmentation}: The speech arrives in real-time, and at each interval (\textit{e.g.,} every 100ms), the detector determines whether the current state of the speaker is SU, Pause, or Gap. That is, we are conducting a ternary segmentation in a sequential manner.
\end{itemize}

\paragraph{Models.} We evaluate a total of four models for end-turn detection.
\begin{itemize}[leftmargin=*,topsep=0pt,parsep=1pt]
\item VAP \citep{ekstedt2022voice}: A popular open-source pretrained turn-taking model, which can also be employed for end-turn detection tasks. We use the pretrained encoder as frozen and train the predictor module of the model from scratch using the OpenETD dataset. We have used the ``comparative'' predictor head, which fits the ETD task.
\item GRU \citep{chung2014empirical}: We construct a lightweight GRU model with about $202\mathrm{K}$ parameters, consisting of a Conv2D frontend ($2$ layers) and a single GRU layer on top of a $40$-dim log--mel input computed on $100\,\mathrm{ms}$ chunks (full input specification in \cref{app:gru_input}), and train it from scratch on the OpenETD dataset.
\item Wav2vec 2.0 \citep{baevski2020wav2vec}: A relatively large transformer-based model with 94M parameters, which achieves high performance in many speech processing tasks. We fine-tune the full model using the OpenETD dataset.
\item SpeculativeETD (Ours): We construct an ETD model using the GRU and Wav2vec 2.0 model following the same settings as described above. Note that, as this method assumes a real-time scenario, we evaluate this method only in real-time audio segmentation scenarios and not in binary classification.
\end{itemize}

\paragraph{Evaluation.} 
For the binary classification task, performance is quantified using Precision, Recall, F1-score, and accuracy metrics. In the real-time segmentation task, we assess the F1-score at 100ms intervals and evaluate segmentation quality using the Intersection over Union (IoU) metric for the average of each of the three classes (Speaking Unit, Gap, Pause).

\paragraph{Datasets for training/evaluation.} In all experiments, we train the models on a mixture of the training splits of the synthetic and real data in the OpenETD dataset; the synthetic split provides controllable pause/gap patterns, while the real split grounds the models in natural conversational speech. For evaluation, we use the held-out test splits of both the synthetic and real datasets, with no overlap with the training or development sets.

\input{table/table2}

\paragraph{Optimization.} We have used the AdamW optimizer \citep{loshchilov2018decoupled} for all tasks and models and trained for 10 epochs. The learning rate has been selected with a random search over the interval $[3 \times 10^{-6}, 3\times 10^{-4}]$, using the validation split of the synthetic OpenETD dataset. The weight decay values are also randomly searched in the range $[0.01,2.00]$. The batch size has been tuned over $\{128,256,512\}$ for GRU, $\{8,16,32\}$ for VAP, and $\{4,8,16\}$ for Wav2vec 2.0. The searched hyperparameters are provided in \cref{appdix:Experiment details}.

\paragraph{Hardware.} 
All experiments have taken place on a single GPU; some experiments have taken place on an NVIDIA L40S, while some have been done on NVIDIA RTX 6000 Ada. To measure the inference latency of ETD models, we used the iPhone 12 mini; the computing took place on the A14 bionic chipset with Hexa-core CPUs.

%% file: table/table2.tex
\begin{table*}[t]
    \small
    \centering
    \resizebox{0.95\linewidth}{!}{%
    \begin{tabular}[t]{l | c c c c | c c c c}
    \toprule
         \multirow{2}{*}{\textbf{Methods}} 
        & \multicolumn{4}{c|}{\textbf{Synthetic data}} 
        & \multicolumn{4}{c}{\textbf{Real data}} \\

        & Pre. & Rec. & F1 & Acc.
        & Pre. & Rec. & F1 & Acc. \\
    \midrule
         VAP {\small \citep{ekstedt2022voice}} 
         & $91.5$ & $93.5$ & $92.1$ & $92.3$ 
            & $59.8$ & $58.7$ & $59.1$ & $69.6$ \\ 
         GRU {\small \citep{chung2014empirical}} & $78.8$ & $77.6$ & $78.1$ & $79.7$ & $52.7$ & $51.4$ & $49.8$ & $69.0$ \\
         wav2vec 2.0 {\small \citep{baevski2020wav2vec}} & $99.3$ & $99.2$ & $99.2$ & $99.3$ & $76.3$ & $74.3$ & $75.2$ & $81.2$ \\
    \bottomrule
\end{tabular}}
\caption{\textbf{Binary classification results.} Performance of various models on the synthetic and real datasets of the OpenETD dataset.}
\label{table:result_binary}
\end{table*}

%% file: sections/6_results.tex
\section{Results}\label{sec:results}

\subsection{Binary classification}

In \cref{table:result_binary}, we report the binary classification results of ETD models. From the table, we observe that the GRU models---which are much lighter than other models---achieve a significantly lower accuracy than other methods, with around 20\%p gap in both synthetic and real datasets. This result motivates us to develop a computationally efficient framework that can achieve both efficiency and accuracy.

\subsection{Real-time audio segmentation}
\input{table/table3}

In \cref{table:result_combined}, we compare the real-time audio segmentation performances of the proposed SpeculativeETD against three baseline methods. The results demonstrate that the SpeculativeETD achieves a performance that is comparable to a larger-scale model, namely Wav2vec 2.0, on the synthetic set. The proposed method also outperforms VAP, which requires much more computation.

\input{figure/figtex/flops_figure}
\input{table/table4}

\subsection{Efficiency of SpeculativeETD}

\paragraph{FLOPs comparison.} 
We evaluate the computational cost of our method in terms of FLOPs using 100 synthetic samples. By leveraging the server-side Wav2vec 2.0 model only when vital, SpeculativeETD achieves comparable performance to Wav2vec 2.0 with substantially lower FLOPs. Moreover, latency remains unaffected as Wav2vec 2.0 runs on a server GPU rather than on-device.

\paragraph{Latency.}
We verify whether the latency is suitable for real-time on-device deployment. We measure the latency via LiteRT\footnote{\url{https://ai.google.dev/edge/litert}} on iPhone mini 12 (iOS 16.3.1). We check model loading, initialization, and execution time, which is shown in~\cref{tab:latency_table}. We averaged three runs of latency measurement, and the execution time was the average of 50 iterations of inference time when the model performed inference every 100 ms on a 5-second speech audio.

Our experiments demonstrate that SpeculativeETD, which utilizes GRU on the on-device side, achieves latency below 1\,ms per 100\,ms interval. As illustrated in \cref{fig:latency_comparison}, the inference latency of Wav2vec 2.0 significantly increases with input length due to the computational overhead of transformer architectures, while GRU latency remains stable. On top of on-device inference, we also measured end-to-end audio-transfer latency over 5G and Wi-Fi; round-trip times stay within $106$--$140\,\mathrm{ms}$ across payload sizes up to $10\,\mathrm{s}$, well inside the $200\,\mathrm{ms}$ turn-taking threshold (see \cref{app:e2e_latency}). These results confirm that SpeculativeETD is an excellent option for serving real-time on-device.

\input{figure/figtex/latency_figure}
\input{table/table5}

%% file: table/table3.tex
\begin{table}[t]
    \small
    \centering
    \resizebox{1\linewidth}{!}{%
    \begin{tabular}{l | c c | c c}
    \toprule
         \multirow{2}{*}{\textbf{Methods}}
         & \multicolumn{2}{c|}{\textbf{Synthetic data}}
         & \multicolumn{2}{c}{\textbf{Real data}} \\
         & F1  & IoU
         & F1  & IoU \\
    \midrule
         VAP {\small \citep{ekstedt2022voice}}
            & 90.6 & 84.8
            & 33.2 & 25.9 \\
         GRU {\small \citep{chung2014empirical}} & 58.0 & 52.2
             & 34.2 & 31.7 \\
         Wav2Vec2 {\small \citep{baevski2020wav2vec}} & \textbf{94.7} & \textbf{90.2}
             & \textbf{58.4} & \textbf{46.2} \\
             \midrule
        SpeculativeETD (Ours) & \underline{94.0} & \underline{88.9}
               & \underline{45.6} & \underline{37.8} \\
    \bottomrule
\end{tabular}}
\caption{\textbf{SpeculativeETD matches Wav2Vec2 on synthetic data and substantially outperforms VAP and GRU baselines on real data.} Real-time 3-class segmentation (Gap/SU/Pause) results on the synthetic and real test sets of the OpenETD dataset. F1 and IoU are macro-averaged across the three states. The best results are marked with \textbf{bold}, and the runner-up is \underline{underlined}. SpeculativeETD retains Wav2Vec2-level synthetic performance while achieving $26.7\times$ fewer W2V calls on real audio.}
\label{table:result_combined}
\end{table}

%% file: figure/figtex/flops_figure.tex
\begin{figure}[t]
    \centering
    \includegraphics[width=0.8\columnwidth]{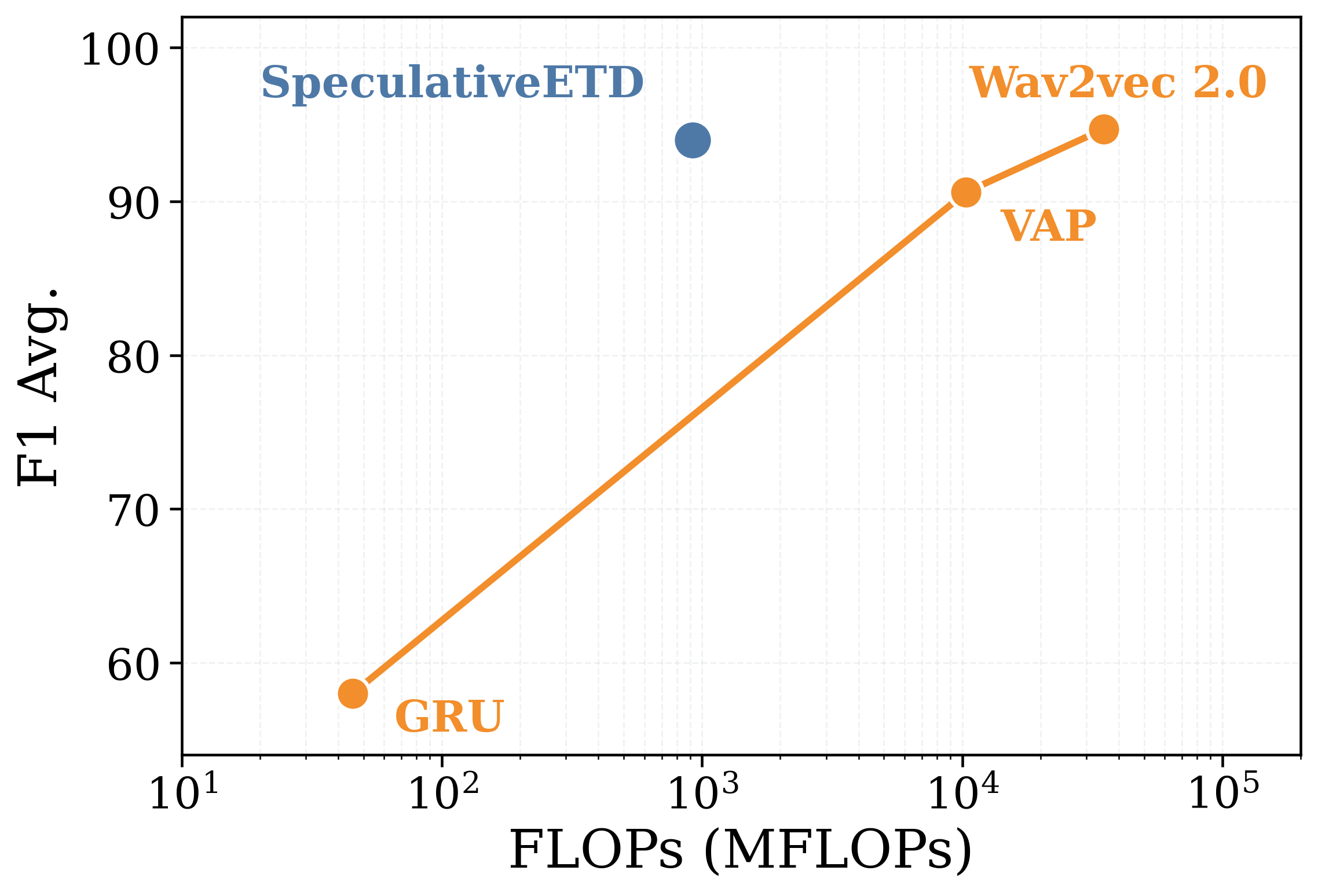}
    \caption{\textbf{FLOPs vs. performance comparison.} Visualization of computational cost (FLOPs) and performance for each model. SpeculativeETD achieves high accuracy with substantially lower computational requirements compared to Wav2vec 2.0.}
    \label{fig:flops_comparison}
\end{figure}

%% file: table/table4.tex
\begin{table}[t]
    \centering
    \resizebox{0.8\columnwidth}{!}{%
        \begin{tabular}{l r}
            \toprule
            \textbf{Methods} & \textbf{Compute (MFLOPs)} \\
            \midrule
            VAP & 10,354.98 \\
            GRU & 45.34 \\
            wav2vec 2.0 & 34,971.68 \\
            \midrule
            SpeculativeETD & {\color{gray}45.34 + 874.30 = }919.64 \\
            \bottomrule
        \end{tabular}
    }
    \caption{\textbf{FLOPs comparison for real-time processing.} Computational cost measured in MFLOPs for processing 100 samples. SpeculativeETD achieves comparable performance to wav2vec 2.0 with 38$\times$ fewer FLOPs.}
    \label{tab:flops_table}
\end{table}

%% file: figure/figtex/latency_figure.tex
\begin{figure}[t]
    \centering
    \includegraphics[width=0.8\columnwidth]{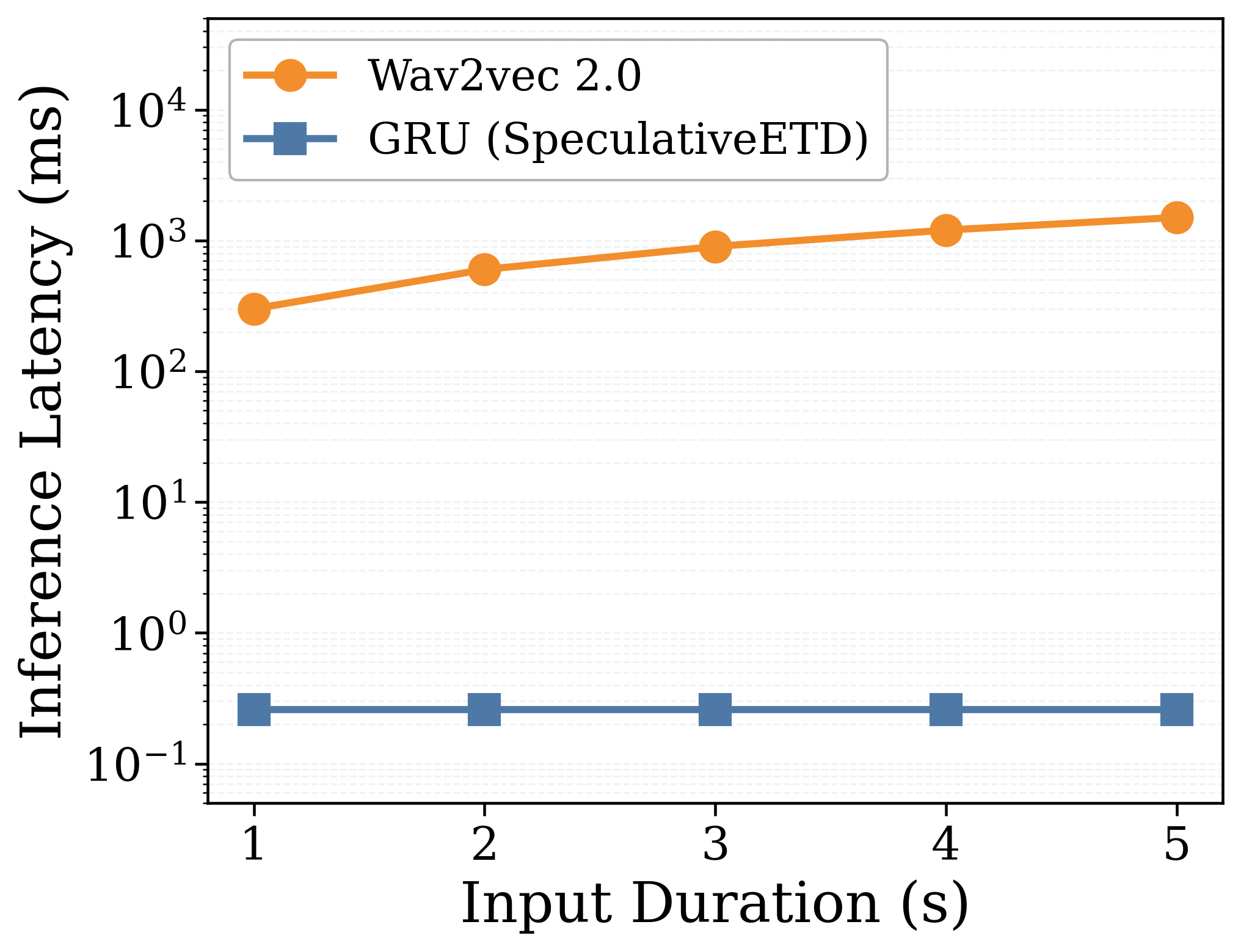}
    \caption{\textbf{Inference latency comparison.} Visualization of inference latency on real-time on-device settings. As input length increases, Wav2vec 2.0 latency grows significantly due to transformer overhead, while GRU maintains consistently low latency, confirming SpeculativeETD's suitability for real-time deployment.}
    \label{fig:latency_comparison}
\end{figure}

%% file: table/table5.tex
\begin{table}[t]
    \centering
    \resizebox{1\columnwidth}{!}{%
        \begin{tabular}{l c c c}
            \toprule
            \multirow{2}{*}{\textbf{Methods}} 
            & \multicolumn{3}{c}{\textbf{Latency (ms)}} \\
            & \textbf{Load} & \textbf{Init} & \textbf{Execute} \\
            \midrule
            wav2vec 2.0 & 874.06 & 17.89 & 1500.32 \\
            GRU (SpeculativeETD) & 1.16 & 3.85 & 0.26 \\
            \bottomrule
        \end{tabular}
    }
    \caption{\textbf{Latency analysis in a real-time on-device setting.} The `Execute'' latency is the average inference time when the model performs inference every 100 ms on a 5-second speech audio. SpeculativeETD with GRU achieves sub-millisecond execution latency.}
    \label{tab:latency_table}
\end{table}

%% file: sections/appendix.tex
\section{Synthetic pause generation: statistical validation}
\label{app:pause_stats}

Our synthetic pipeline does not insert pauses at random positions. Instead, we detect micro-hesitations naturally produced by the TTS model \citep{braunschweiler2011automatic, hwang2023pausespeech, yang2024frame} and extend their duration using an Erlang distribution fitted to real pause durations.

\paragraph{Duration.} \Cref{tab:pause_stats} reports Kolmogorov--Smirnov (KS) two-sample statistics and Cohen's~$d$ comparing synthetic and real gap/pause durations. The KS statistic for gap duration is only $0.083$ (Cohen's $d=0.12$), indicating that the Erlang-fitted generator closely matches real duration distributions.

\paragraph{Position.} In real conversational data, both pause and gap positions within an utterance are close to uniform (KS vs.\ Uniform$(0,1)$: pause $0.091$, gap $0.094$), reflecting the inherent unpredictability of pause placement. Synthetic positions show a larger divergence (KS $=0.238$ for both pause and gap). We acknowledge this as a limitation; however, exact position replication is inherently difficult given that real positions themselves are near-uniform.

\paragraph{Augmentation value.} The synthetic data therefore functions as augmentation rather than exact replication. \Cref{tab:training_mode_v2} shows that training SpeculativeETD on the mixture consistently outperforms training on either split alone, confirming the practical value of the synthetic set.

\input{table/appendix/pause_stats_v2}
\input{table/appendix/training_mode_v2}

\section{On-device GRU input specification}
\label{app:gru_input}

We detail the input pipeline of the on-device GRU used in SpeculativeETD.

\paragraph{Mel-spectrogram input.} Each input chunk corresponds to $100\,\mathrm{ms}$ of audio sampled at $16\,\mathrm{kHz}$. We extract a $40$-dim log--mel spectrogram with a $25\,\mathrm{ms}$ window and a $10\,\mathrm{ms}$ hop, yielding $10$ frames per chunk, followed by per-utterance z-score normalization. The resulting chunk tensor has shape $(40, 10)$.

\paragraph{Conv2D frontend.} Two Conv2D layers downsample the chunk: $\mathrm{Conv}(1\!\to\!16, k{=}5, s{=}2, p{=}2)\,{\rightarrow}\,\mathrm{ReLU}\,{\rightarrow}\,\mathrm{Conv}(16\!\to\!32, k{=}3, s{=}2, p{=}1)\,{\rightarrow}\,\mathrm{ReLU}$, producing a $960$-dim per-chunk feature vector after flattening.

\paragraph{GRU backbone.} A single-layer GRU (hidden size $64$) consumes the chunk features autoregressively, and a linear head outputs the binary (SU vs.\ non-SU) logits. The entire on-device model has approximately $202\mathrm{K}$ parameters.

\paragraph{Filler word vocabulary.} During synthetic data generation (\cref{ssec:synthetic}), we inject filler words drawn from the following vocabulary: ``uh'', ``um'', ``ah'', ``er'', ``hmm'', ``mhm'', ``uh huh'', ``like'', ``you know'', ``so'', ``actually um'', and ``basically um''. These cover the common English hesitation markers observed in conversational speech.

\section{Human validation of pause/gap labels}
\label{app:human_validation}

To quantify label noise introduced by the automatic diarization + 200\,ms rule, we performed a human validation study. Three annotators each judged $96$ randomly sampled real clips ($50$ Pause, $46$ Gap). Each clip was presented with $2\,\mathrm{s}$ of audio before the silence and $3\,\mathrm{s}$ after, and annotators reported (i) whether the automatic label was correct and (ii) the diarization segmentation quality on a $1$--$5$ scale.

\Cref{tab:human_validation} summarizes the results. Pause labels are highly reliable (human--auto agreement $94.0\%$), while Gap labels are moderately noisier (76.1\%) because speaker-change boundaries are intrinsically harder to localize. Overall diarization quality averages $4.17/5.0$, supporting the use of the automatic pipeline. The $200\,\mathrm{ms}$ threshold follows the widely adopted convention in turn-taking research~\citep{doi:10.1073/pnas.0903616106, HELDNER2010555}.

\input{table/appendix/human_validation}

\section{Domain analysis: synthetic vs.\ real}
\label{app:domain_analysis}

To characterize the mismatch between synthetic and real conversations, we analyzed audio-level and speaker-level statistics (\cref{tab:domain_analysis}). Signal-to-noise ratio (SNR) is computed from the raw waveform, Word-Per-Second (WPS) from Whisper-tiny transcripts, gender is derived from metadata, and speaker count reflects the distinct voice identities present.

The two largest domain gaps are SNR ($39.7\,\mathrm{dB}$ vs.\ $10.5\,\mathrm{dB}$) and speaking-rate variability ($\pm 0.77$ vs.\ $\pm 3.30$). Real data also exhibits a richer emotion distribution (with a substantially higher share of sad speech) and spans five English accents, whereas synthetic data uses only two TTS voices (EN-US and EN-BR); see \cref{tab:domain_analysis}.

\input{table/appendix/domain_analysis}

\section{End-to-end audio transfer latency}
\label{app:e2e_latency}

Although SpeculativeETD operates asynchronously---the on-device GRU continues detecting silence while the server classifies in the background---excessively long transfer times could still delay the classification decision. We therefore measured real audio-transfer round-trip time (RTT) from an iPhone 12 mini to a remote server via WebSocket, without model inference, across six payload sizes ($100\,\mathrm{ms}$--$10\,\mathrm{s}$ of $16\,\mathrm{kHz}$ audio; $50$ trials each).

\Cref{tab:e2e_latency} reports the results. On 5G, RTT ranges from $106$ to $116\,\mathrm{ms}$; on Wi-Fi, $98$ to $140\,\mathrm{ms}$---both comfortably inside the $200\,\mathrm{ms}$ turn-taking threshold. RTT is dominated by fixed network latency rather than payload size; a $100\times$ payload increase ($3.1\,\mathrm{KB} \to 312.5\,\mathrm{KB}$) adds only about $10\,\mathrm{ms}$ on 5G.

\input{table/appendix/e2e_latency}

\section{Licensing of data sources}
\label{app:licenses}

\Cref{tab:licenses} summarizes the licenses of the data sources used to build OpenETD. We release the dataset and processing code in a manner consistent with each underlying license: the MultiWOZ-derived synthetic data inherits Apache~2.0; YouTube audio is not redistributed (we share URLs and download scripts only); Buckeye Corpus is used under its Academic License.

\input{table/appendix/license_table}

\section{Data quality improvement}

\paragraph{Segment Filtering.}
We applied a structured segmentation process to better capture conversational dynamics. Consecutive speech fragments from the same speaker were merged into a Speaking Unit (SU) if separated by less than 200\,ms of silence. Silences longer than $200\,\mathrm{ms}$ were labeled Pause (if within the same speaker) or Gap (if between different speakers), based on established turn-taking criteria~\citep{doi:10.1073/pnas.0903616106, HELDNER2010555}. Using these labels, we grouped continuous SUs into segments, where each segment represents a stretch of speech bounded by either a Pause or a Gap. To focus on turn-relevant phenomena, we filtered out segments that did not contain any Pause or Gap, retaining only those that reflect actual hesitation or speaker transitions crucial for end-turn detection.

\input{table/appendix/realdata_language}
\paragraph{Language Filtering.}
To ensure the quality and consistency of the real-world speech data, we applied a language filtering step using the Whisper model \citep{radford2022robustspeechrecognitionlargescale}. For each audio file, we used Whisper's language detection capability to identify the spoken language. Since our focus is on English-based dialogue systems, we retained only the samples identified as English and discarded those in other languages. \cref{tab:lang_dist} summarizes the top 10 most frequent languages and the aggregate of other low-resource languages. Out of a total of 11{,}127 collected files, 8{,}022 were classified as English and included in the final dataset. We additionally conducted a human evaluation of language identification performance by Whisper, which is recognized well as English. We checked all the data, and the accuracy of identification was $99.07\%$.

\section{Experiment details}
\label{appdix:Experiment details}
\input{table/appendix/hyper-parameter}

We detail the hyperparameter settings used for training each model in our experiments. All models were trained for 10 epochs. The VAP model was configured with a batch size of 32 and a learning rate of $6 \times 10^{-6}$. For the GRU model, we used a larger batch size of 256 and set the learning rate to $3 \times 10^{-4}$. In the case of Wav2vec 2.0, we employed a smaller batch size of 8 with a learning rate of $5 \times 10^{-6}$.

%% file: table/appendix/pause_stats_v2.tex
\begin{table}[t]
\small
\centering
\resizebox{\columnwidth}{!}{%
\begin{tabular}{l c c}
\toprule
\textbf{Comparison} & \textbf{KS stat} & \textbf{Cohen's $d$} \\
\midrule
\multicolumn{3}{c}{\textit{Duration}} \\
\midrule
Syn vs.\ Real gap duration  & 0.083 & 0.12 \\
Syn vs.\ Real pause duration & 0.092 & 0.15 \\
\midrule
\multicolumn{3}{c}{\textit{Position within utterance}} \\
\midrule
Real pause vs.\ Uniform$(0,1)$ & 0.091 & -- \\
Real gap vs.\ Uniform$(0,1)$   & 0.094 & -- \\
Syn vs.\ Real gap (position)   & 0.238 & 0.23 \\
Syn vs.\ Real pause (position) & 0.238 & 0.27 \\
\bottomrule
\end{tabular}%
}
\caption{\textbf{Synthetic pauses match real duration statistics but not exact positions.} Kolmogorov--Smirnov (KS) two-sample test and Cohen's $d$ comparing synthetic and real pause/gap distributions. Duration statistics are nearly identical (KS$=0.083$), confirming that the Erlang-fitted generator captures real pause lengths. Real-data positions are close to uniform within an utterance, so exact position replication is inherently difficult; synthetic data therefore serves as augmentation rather than perfect replication.}
\label{tab:pause_stats}
\end{table}

%% file: table/appendix/training_mode_v2.tex
\begin{table}[t]
\small
\centering
\resizebox{\columnwidth}{!}{%
\begin{tabular}{l c c c c}
\toprule
\textbf{Training mode} & \textbf{Real F1} & \textbf{Real IoU} & \textbf{$\Delta$F1} & \textbf{$\Delta$IoU} \\
\midrule
Mix (syn + real) & \textbf{45.6} & \textbf{37.8} & --   & --   \\
Real only        & 43.1          & 36.3          & $-2.5$ & $-1.5$ \\
Synthetic only   & 44.0          & 36.7          & $-1.6$ & $-1.1$ \\
\bottomrule
\end{tabular}%
}
\caption{\textbf{Mixed training consistently improves real-data performance, confirming synthetic data as effective augmentation.} SpeculativeETD real-time segmentation performance on the real test set when trained on synthetic only, real only, or the mixture. Mixed training achieves the best F1 and IoU; $\Delta$ denotes degradation relative to the Mix row.}
\label{tab:training_mode_v2}
\end{table}

%% file: table/appendix/human_validation.tex
\begin{table}[t]
\small
\centering
\resizebox{\columnwidth}{!}{%
\begin{tabular}{l c c c}
\toprule
\textbf{Metric} & \textbf{Pause ($n{=}50$)} & \textbf{Gap ($n{=}46$)} & \textbf{Overall ($n{=}96$)} \\
\midrule
Human--Auto agreement            & 94.0\% & 76.1\% & 85.4\% \\
Diarization quality (1--5)       & $4.29\pm1.00$ & $4.04\pm1.27$ & $4.17\pm1.14$ \\
\bottomrule
\end{tabular}%
}
\caption{\textbf{Human validation confirms reliable pause labels and moderate gap-label noise.} Three annotators judged label correctness and diarization segmentation quality on 96 randomly sampled real clips (2\,s before + silence + 3\,s after). Pause labels show $94\%$ human agreement, while Gap boundaries are harder to pinpoint (76.1\%) due to ambiguous speaker-change boundaries. Diarization quality averages $4.17/5$, indicating the automatic pipeline generally produces clean segmentations.}
\label{tab:human_validation}
\end{table}

%% file: table/appendix/domain_analysis.tex
\begin{table}[t]
\small
\centering
\begin{subtable}{\columnwidth}
\centering
\resizebox{\columnwidth}{!}{%
\begin{tabular}{l c c}
\toprule
\textbf{Dimension} & \textbf{Synthetic} & \textbf{Real} \\
\midrule
Signal-to-noise ratio (dB) & $39.7 \pm 4.2$  & $10.5 \pm 13.7$ \\
Word per second            & $1.57 \pm 0.77$ & $2.87 \pm 3.30$ \\
Gender (Female \%)         & 66.2            & 39.4 \\
Speaker count              & 3 TTS voices    & $\sim$200 \\
\bottomrule
\end{tabular}%
}
\subcaption{Audio / speaker statistics.}
\end{subtable}

\vspace{4pt}

\begin{subtable}{\columnwidth}
\centering
\resizebox{\columnwidth}{!}{%
\begin{tabular}{l c c c c c | c c c c c}
\toprule
& \multicolumn{5}{c|}{\textbf{Emotion (\%)}} & \multicolumn{5}{c}{\textbf{Accent, Real (\%)}} \\
& Happy & Sad & Angry & Disgust & Other & US & England & Canada & Indian & Australia \\
\midrule
Real      & 33 & 33 & 11 & 5  & 18 & 44.0 & 16.6 & 16.6 & 14.6 & 8.2 \\
Synthetic & 30 &  8 & 19 & 30 & 13 & \multicolumn{5}{c}{Two TTS voices (EN-US, EN-BR)} \\
\bottomrule
\end{tabular}%
}
\subcaption{Emotion distribution and real-data accent distribution.}
\end{subtable}

\caption{\textbf{Domain gaps between synthetic and real data: lower SNR, higher speaking-rate variability, richer emotion and accent diversity.} (a) Audio-level and speaker-level statistics; SNR is computed from the raw waveform, speaking rate from Whisper-tiny transcripts, gender/speaker from metadata. (b) Emotion distribution (wav2vec2-based classifier) and English-accent distribution on the real side (\texttt{dima806/speech-accent-classification}); the synthetic set only contains two TTS accent voices. These gaps motivate real-data training and highlight limitations that synthetic augmentation cannot fully close.}
\label{tab:domain_analysis}
\end{table}

%% file: table/appendix/e2e_latency.tex
\begin{table}[t]
\small
\centering
\resizebox{\columnwidth}{!}{%
\begin{tabular}{l l c c c}
\toprule
\textbf{Audio} & \textbf{Payload} & \textbf{Mean RTT} & \textbf{P50} & \textbf{P95} \\
\midrule
\multicolumn{5}{c}{\textit{5G network}} \\
\midrule
100\,ms & 3.1\,KB   & 106.0\,ms & 104.0\,ms & 115.0\,ms \\
500\,ms & 15.6\,KB  & 106.4\,ms & 104.0\,ms & 119.0\,ms \\
1\,s    & 31.3\,KB  & 106.4\,ms & 104.0\,ms & 119.0\,ms \\
3\,s    & 93.8\,KB  & 107.2\,ms & 103.0\,ms & 110.0\,ms \\
6\,s    & 187.5\,KB & 109.9\,ms & 108.0\,ms & 128.0\,ms \\
10\,s   & 312.5\,KB & 116.2\,ms & 114.0\,ms & 129.0\,ms \\
\midrule
\multicolumn{5}{c}{\textit{Wi-Fi}} \\
\midrule
100\,ms & 3.1\,KB   & 100.2\,ms &  95.0\,ms & 131.0\,ms \\
500\,ms & 15.6\,KB  & 101.6\,ms &  94.0\,ms & 142.0\,ms \\
1\,s    & 31.3\,KB  & 107.9\,ms &  99.0\,ms & 178.0\,ms \\
3\,s    & 93.8\,KB  &  98.8\,ms &  94.0\,ms & 121.0\,ms \\
6\,s    & 187.5\,KB & 107.0\,ms & 100.0\,ms & 128.0\,ms \\
10\,s   & 312.5\,KB & 139.7\,ms & 115.0\,ms & 225.0\,ms \\
\bottomrule
\end{tabular}%
}
\caption{\textbf{End-to-end audio transfer stays well within the 200\,ms turn-taking threshold across 5G and Wi-Fi.} Round-trip time (RTT) measured from iPhone 12 mini to a remote server via WebSocket without model inference, across six payload sizes (50 trials each). RTT is dominated by fixed network latency rather than payload size: a $100\times$ payload increase adds only $\sim$10\,ms on 5G.}
\label{tab:e2e_latency}
\end{table}

%% file: table/appendix/license_table.tex
\begin{table}[h]
\small
\centering
\resizebox{\columnwidth}{!}{%
\begin{tabular}{l l}
\toprule
\textbf{Dataset / Source} & \textbf{License} \\
\midrule
MultiWOZ \citep{budzianowski-etal-2018-multiwoz} & Apache 2.0 \\
Google Cloud Text-to-Speech     & Google Cloud Terms of Service \\
YouTube audio                   & URLs + download scripts (not redistributed) \\
Buckeye Corpus \citep{Pitt2005TheBC} & Academic License \\
\bottomrule
\end{tabular}%
}
\caption{\textbf{Licenses of the data sources used to build OpenETD.} We release the MultiWOZ-derived synthetic data under a compatible license, redistribute only scripts and URLs for YouTube content, and respect the Buckeye Academic License in all downstream use.}
\label{tab:licenses}
\end{table}

%% file: table/appendix/realdata_language.tex
\begin{table}[h]
\centering
\resizebox{1\columnwidth}{!}{
\begin{tabular}{l r r c}
\toprule
\textbf{Language} & \textbf{Code} & \textbf{\# Files} & \textbf{Included} \\
\midrule
English & en & 8,022 & \checkmark \\
\midrule
Hindi & hi & 674 & \ding{55} \\
Romanian & ro & 334 & \ding{55} \\
Urdu & ur & 315 & \ding{55} \\
Spanish & es & 258 & \ding{55} \\
Arabic & ar & 179 & \ding{55} \\
Swahili & sw & 151 & \ding{55} \\
Korean & ko & 43 & \ding{55} \\
Portuguese & pt & 41 & \ding{55} \\
French & fr & 25 & \ding{55} \\
Others (40+ languages) & — & 1,055 & \ding{55} \\
\bottomrule
\end{tabular}
}
\caption{\textbf{Language distribution before filtering.} Top-10 most frequent languages in the collected data. Only English files were retained in the final dataset.}
\label{tab:lang_dist}
\end{table}

%% file: table/appendix/hyper-parameter.tex
\begin{table}[h]
\centering
\resizebox{1\linewidth}{!}{
\begin{tabular}{lccc}
\toprule
\multirow{2}{*}{Model} 
 & \multicolumn{3}{c}{\textbf{Hyper-parameter}} \\
\cmidrule(r){2-4}
& epochs & batch size & lr     \\
\midrule
VAP & 10 & 32 & $6 \times 10^{-6}$   \\
GRU & 10 & 256 & $3 \times 10^{-4}$   \\
Wav2vec 2.0 & 10 & 8 & $5 \times 10^{-6}$   \\
\bottomrule
\end{tabular}
}
\caption{\textbf{Hyper-parameter settings.} Training configuration used for each model in our experiments.}
\label{parametersetting}
\end{table}

%% file: custom.bib
@inproceedings{song2023llm,
  title={Llm-planner: Few-shot grounded planning for embodied agents with large language models},
  author={Song, Chan Hee and Wu, Jiaman and Washington, Clayton and Sadler, Brian M and Chao, Wei-Lun and Su, Yu},
  booktitle={Proceedings of the IEEE/CVF International Conference on Computer Vision},
  pages={2998--3009},
  year={2023}
}

@inproceedings{shao2024assisting,
  title={Assisting in Writing Wikipedia-like Articles From Scratch with Large Language Models},
  author={Shao, Yijia and Jiang, Yucheng and Kanell, Theodore and Xu, Peter and Khattab, Omar and Lam, Monica},
  booktitle={Proceedings of the 2024 Conference of the North American Chapter of the Association for Computational Linguistics: Human Language Technologies (Volume 1: Long Papers)},
  pages={6252--6278},
  year={2024}
}

@inproceedings{zhang-etal-2024-agent,
    title = "Agent-Pro: Learning to Evolve via Policy-Level Reflection and Optimization",
    author = "Zhang, Wenqi  and
      Tang, Ke  and
      Wu, Hai  and
      Wang, Mengna  and
      Shen, Yongliang  and
      Hou, Guiyang  and
      Tan, Zeqi  and
      Li, Peng  and
      Zhuang, Yueting  and
      Lu, Weiming",
    editor = "Ku, Lun-Wei  and
      Martins, Andre  and
      Srikumar, Vivek",
    booktitle = "Proceedings of the 62nd Annual Meeting of the Association for Computational Linguistics (Volume 1: Long Papers)",
    month = aug,
    year = "2024",
    address = "Bangkok, Thailand",
    publisher = "Association for Computational Linguistics",
    url = "https://aclanthology.org/2024.acl-long.292/",
    doi = "10.18653/v1/2024.acl-long.292",
    pages = "5348--5375"
}

@inproceedings{
xie2024can,
title={Can Large Language Model Agents Simulate Human Trust Behavior?},
author={Chengxing Xie and Canyu Chen and Feiran Jia and Ziyu Ye and Shiyang Lai and Kai Shu and Jindong Gu and Adel Bibi and Ziniu Hu and David Jurgens and James Evans and Philip Torr and Bernard Ghanem and Guohao Li},
booktitle={The Thirty-eighth Annual Conference on Neural Information Processing Systems},
year={2024},
url={https://openreview.net/forum?id=CeOwahuQic}
}

@article{mitsui2023towards,
  title={Towards human-like spoken dialogue generation between AI agents from written dialogue},
  author={Mitsui, Kentaro and Hono, Yukiya and Sawada, Kei},
  journal={arXiv preprint arXiv:2310.01088},
  year={2023}
}

@inproceedings{yan-etal-2024-talk,
    title = "Talk With Human-like Agents: Empathetic Dialogue Through Perceptible Acoustic Reception and Reaction",
    author = "Yan, Haoqiu  and
      Zhu, Yongxin  and
      Zheng, Kai  and
      Liu, Bing  and
      Cao, Haoyu  and
      Jiang, Deqiang  and
      Xu, Linli",
    editor = "Ku, Lun-Wei  and
      Martins, Andre  and
      Srikumar, Vivek",
    booktitle = "Proceedings of the 62nd Annual Meeting of the Association for Computational Linguistics (Volume 1: Long Papers)",
    month = aug,
    year = "2024",
    address = "Bangkok, Thailand",
    publisher = "Association for Computational Linguistics",
    url = "https://aclanthology.org/2024.acl-long.801/",
    doi = "10.18653/v1/2024.acl-long.801",
    pages = "15009--15022"
}

@article{ma2024language,
  title={Language Model Can Listen While Speaking},
  author={Ma, Ziyang and Song, Yakun and Du, Chenpeng and Cong, Jian and Chen, Zhuo and Wang, Yuping and Wang, Yuxuan and Chen, Xie},
  journal={arXiv preprint arXiv:2408.02622},
  year={2024}
}

@inproceedings{veluri2024beyond,
  title={Beyond Turn-Based Interfaces: Synchronous LLMs as Full-Duplex Dialogue Agents},
  author={Veluri, Bandhav and Peloquin, Benjamin and Yu, Bokai and Gong, Hongyu and Gollakota, Shyamnath},
  booktitle={Proceedings of the 2024 Conference on Empirical Methods in Natural Language Processing},
  pages={21390--21402},
  year={2024}
}

@inproceedings{lin2022duplex,
  title={Duplex conversation: Towards human-like interaction in spoken dialogue systems},
  author={Lin, Ting-En and Wu, Yuchuan and Huang, Fei and Si, Luo and Sun, Jian and Li, Yongbin},
  booktitle={Proceedings of the 28th ACM SIGKDD Conference on Knowledge Discovery and Data Mining},
  pages={3299--3308},
  year={2022}
}

@inproceedings{umair2024large,
  title={Large Language Models Know What To Say But Not When To Speak},
  author={Umair, Muhammad and Sarathy, Vasanth and Ruiter, Jan},
  booktitle={Findings of the Association for Computational Linguistics: EMNLP 2024},
  pages={15503--15514},
  year={2024}
}

@article{chang2022turn,
  title={Turn-taking prediction for natural conversational speech},
  author={Chang, Shuo-yiin and Li, Bo and Sainath, Tara N and Zhang, Chao and Strohman, Trevor and Liang, Qiao and He, Yanzhang},
  journal={arXiv preprint arXiv:2208.13321},
  year={2022}
}

@inproceedings{cieri-etal-2004-fisher,
    title = "The Fisher Corpus: a Resource for the Next Generations of Speech-to-Text",
    author = "Cieri, Christopher  and
      Miller, David  and
      Walker, Kevin",
    editor = "Lino, Maria Teresa  and
      Xavier, Maria Francisca  and
      Ferreira, F{\'a}tima  and
      Costa, Rute  and
      Silva, Raquel",
    booktitle = "Proceedings of the Fourth International Conference on Language Resources and Evaluation ({LREC}`04)",
    month = may,
    year = "2004",
    address = "Lisbon, Portugal",
    publisher = "European Language Resources Association (ELRA)",
    url = "https://aclanthology.org/L04-1500/"
}

@article{10.1155/2023/5870630,
author = {Qi, Xuan and Sun, Zegang and Mei, Xue and Chellali, Ryad and Yi, Yugen},
title = {A Lightweight Binarized Convolutional Neural Network Model for Small Memory and Low-Cost Mobile Devices},
year = {2023},
issue_date = {2023},
publisher = {IOS Press},
address = {NLD},
volume = {2023},
issn = {1574-017X},
url = {https://doi.org/10.1155/2023/5870630},
doi = {10.1155/2023/5870630},
journal = {Mob. Inf. Syst.},
month = jan,
numpages = {11}
}

@article{10.4018/IJSWIS.330015,
author = {Wu, YiHeng and Chen, JianXin},
title = {A Lightweight Real-Time System for Object Detection in Enterprise Information Systems for Frequency-Based Feature Separation},
year = {2023},
issue_date = {Jun 2023},
publisher = {IGI Global},
address = {USA},
volume = {19},
number = {1},
issn = {1552-6283},
url = {https://doi.org/10.4018/IJSWIS.330015},
doi = {10.4018/IJSWIS.330015},
journal = {Int. J. Semant. Web Inf. Syst.},
month = sep,
pages = {1–18},
numpages = {18}
}

@inproceedings{chung2014empirical,
  title={Empirical evaluation of gated recurrent neural networks on sequence modeling},
  author={Chung, Junyoung and Gulcehre, Caglar and Cho, Kyunghyun and Bengio, Yoshua},
  booktitle={NIPS 2014 Workshop on Deep Learning, December 2014},
  year={2014}
}

@article{baevski2020wav2vec,
  title={wav2vec 2.0: A framework for self-supervised learning of speech representations},
  author={Baevski, Alexei and Zhou, Yuhao and Mohamed, Abdelrahman and Auli, Michael},
  journal={Advances in neural information processing systems},
  volume={33},
  pages={12449--12460},
  year={2020}
}

@inproceedings{budzianowski-etal-2018-multiwoz,
    title = "{M}ulti{WOZ} - A Large-Scale Multi-Domain {W}izard-of-{O}z Dataset for Task-Oriented Dialogue Modelling",
    author = "Budzianowski, Pawe{\l}  and
      Wen, Tsung-Hsien  and
      Tseng, Bo-Hsiang  and
      Casanueva, I{\~n}igo  and
      Ultes, Stefan  and
      Ramadan, Osman  and
      Ga{\v{s}}i{\'c}, Milica",
    editor = "Riloff, Ellen  and
      Chiang, David  and
      Hockenmaier, Julia  and
      Tsujii, Jun{'}ichi",
    booktitle = "Proceedings of the 2018 Conference on Empirical Methods in Natural Language Processing",
    month = oct # "-" # nov,
    year = "2018",
    address = "Brussels, Belgium",
    publisher = "Association for Computational Linguistics",
    url = "https://aclanthology.org/D18-1547/",
    doi = "10.18653/v1/D18-1547",
    pages = "5016--5026"
}

@inproceedings{ekstedt-skantze-2020-turngpt,
    title = "{T}urn{GPT}: a Transformer-based Language Model for Predicting Turn-taking in Spoken Dialog",
    author = "Ekstedt, Erik  and
      Skantze, Gabriel",
    editor = "Cohn, Trevor  and
      He, Yulan  and
      Liu, Yang",
    booktitle = "Findings of the Association for Computational Linguistics: EMNLP 2020",
    month = nov,
    year = "2020",
    address = "Online",
    publisher = "Association for Computational Linguistics",
    url = "https://aclanthology.org/2020.findings-emnlp.268/",
    doi = "10.18653/v1/2020.findings-emnlp.268",
    pages = "2981--2990"
}

@misc{ghosh2022spanclassificationstructuredinformation,
      title={Span Classification with Structured Information for Disfluency Detection in Spoken Utterances}, 
      author={Sreyan Ghosh and Sonal Kumar and Yaman Kumar Singla and Rajiv Ratn Shah and S. Umesh},
      year={2022},
      eprint={2203.16028},
      archivePrefix={arXiv},
      primaryClass={cs.CL},
      url={https://arxiv.org/abs/2203.16028}, 
}

@misc{zhou2024yolostutterendtoendregionwisespeech,
      title={YOLO-Stutter: End-to-end Region-Wise Speech Dysfluency Detection}, 
      author={Xuanru Zhou and Anshul Kashyap and Steve Li and Ayati Sharma and Brittany Morin and David Baquirin and Jet Vonk and Zoe Ezzes and Zachary Miller and Maria Luisa Gorno Tempini and Jiachen Lian and Gopala Krishna Anumanchipalli},
      year={2024},
      eprint={2408.15297},
      archivePrefix={arXiv},
      primaryClass={eess.AS},
      url={https://arxiv.org/abs/2408.15297}, 
}

@misc{lou2020endtoendspeechrecognitiondisfluency,
      title={End-to-End Speech Recognition and Disfluency Removal}, 
      author={Paria Jamshid Lou and Mark Johnson},
      year={2020},
      eprint={2009.10298},
      archivePrefix={arXiv},
      primaryClass={eess.AS},
      url={https://arxiv.org/abs/2009.10298}, 
}

@inproceedings{leviathan2023fast,
  title={Fast inference from transformers via speculative decoding},
  author={Leviathan, Yaniv and Kalman, Matan and Matias, Yossi},
  booktitle={International Conference on Machine Learning},
  pages={19274--19286},
  year={2023},
  organization={PMLR}
}

@inproceedings{raux-eskenazi-2009-finite,
    title = "A Finite-State Turn-Taking Model for Spoken Dialog Systems",
    author = "Raux, Antoine  and
      Eskenazi, Maxine",
    editor = "Ostendorf, Mari  and
      Collins, Michael  and
      Narayanan, Shri  and
      Oard, Douglas W.  and
      Vanderwende, Lucy",
    booktitle = "Proceedings of Human Language Technologies: The 2009 Annual Conference of the North {A}merican Chapter of the Association for Computational Linguistics",
    month = jun,
    year = "2009",
    address = "Boulder, Colorado",
    publisher = "Association for Computational Linguistics",
    url = "https://aclanthology.org/N09-1071/",
    pages = "629--637"
}

@article{Levinson2015Timing,
  author    = {Stephen C. Levinson and Francisco Torreira},
  title     = {Timing in turn-taking and its implications for processing models of language},
  journal   = {Frontiers in Psychology},
  volume    = {6},
  pages     = {731},
  year      = {2015},
  doi       = {10.3389/fpsyg.2015.00731},
  publisher = {Frontiers Media SA}
}

@article{gravano2011turn,
author = {Gravano, Agustín and Hirschberg, Julia},
year = {2011},
month = {07},
pages = {601-634},
title = {Turn-taking cues in task-oriented dialogue},
volume = {25},
journal = {Computer Speech \& Language},
doi = {10.1016/j.csl.2010.10.003}
}

@article{ten2005temporal,
  title={On temporal aspects of turn taking in conversational dialogues},
  author={Ten Bosch, Louis and Oostdijk, Nelleke and Boves, Lou},
  journal={Speech Communication},
  volume={47},
  number={1-2},
  pages={80--86},
  year={2005},
  publisher={Elsevier}
}

@inproceedings{
loshchilov2018decoupled,
title={Decoupled Weight Decay Regularization},
author={Loshchilov, Ilya and Hutter, Frank},
booktitle={ICLR},
year={2019},
url={https://openreview.net/forum?id=Bkg6RiCqY7},
}

@article{ekstedt2022voice,
  title={Voice activity projection: Self-supervised learning of turn-taking events},
  author={Ekstedt, Erik and Skantze, Gabriel},
  journal={arXiv preprint arXiv:2205.09812},
  year={2022}
}

@article{rubenstein2023audiopalm,
  title={Audiopalm: A large language model that can speak and listen},
  author={Rubenstein, Paul K and Asawaroengchai, Chulayuth and Nguyen, Duc Dung and Bapna, Ankur and Borsos, Zal{\'a}n and Quitry, F{\'e}lix de Chaumont and Chen, Peter and Badawy, Dalia El and Han, Wei and Kharitonov, Eugene and others},
  journal={arXiv preprint arXiv:2306.12925},
  year={2023}
}

@misc{defossez2024moshi,
      title={Moshi: a speech-text foundation model for real-time dialogue}, 
      author={Alexandre Défossez and Laurent Mazaré and Manu Orsini and Amélie Royer and Patrick Pérez and Hervé Jégou and Edouard Grave and Neil Zeghidour},
      year={2024},
      eprint={2410.00037},
      archivePrefix={arXiv},
      primaryClass={eess.AS},
      url={https://arxiv.org/abs/2410.00037}, 
}

@misc{nguyen2022generativespokendialoguelanguage,
      title={Generative Spoken Dialogue Language Modeling}, 
      author={Tu Anh Nguyen and Eugene Kharitonov and Jade Copet and Yossi Adi and Wei-Ning Hsu and Ali Elkahky and Paden Tomasello and Robin Algayres and Benoit Sagot and Abdelrahman Mohamed and Emmanuel Dupoux},
      year={2022},
      eprint={2203.16502},
      archivePrefix={arXiv},
      primaryClass={cs.CL},
      url={https://arxiv.org/abs/2203.16502}, 
}

@misc{li2022ispeakpredictinginitiation,
      title={When can I Speak? Predicting initiation points for spoken dialogue agents}, 
      author={Siyan Li and Ashwin Paranjape and Christopher D. Manning},
      year={2022},
      eprint={2208.03812},
      archivePrefix={arXiv},
      primaryClass={cs.CL},
      url={https://arxiv.org/abs/2208.03812}, 
}

@article{HELDNER2010555,
title = {Pauses, gaps and overlaps in conversations},
journal = {Journal of Phonetics},
volume = {38},
number = {4},
pages = {555-568},
year = {2010},
issn = {0095-4470},
doi = {https://doi.org/10.1016/j.wocn.2010.08.002},
url = {https://www.sciencedirect.com/science/article/pii/S0095447010000628},
author = {Mattias Heldner and Jens Edlund}
}

@article{
doi:10.1073/pnas.0903616106,
author = {Tanya Stivers  and N. J. Enfield  and Penelope Brown  and Christina Englert  and Makoto Hayashi  and Trine Heinemann  and Gertie Hoymann  and Federico Rossano  and Jan Peter de Ruiter  and Kyung-Eun Yoon  and Stephen C. Levinson },
title = {Universals and cultural variation in turn-taking in conversation},
journal = {Proceedings of the National Academy of Sciences},
volume = {106},
number = {26},
pages = {10587-10592},
year = {2009},
doi = {10.1073/pnas.0903616106},
URL = {https://www.pnas.org/doi/abs/10.1073/pnas.0903616106},
eprint = {https://www.pnas.org/doi/pdf/10.1073/pnas.0903616106}}

@inproceedings{Bredin23,
  author={Hervé Bredin},
  title={{pyannote.audio 2.1 speaker diarization pipeline: principle, benchmark, and recipe}},
  year=2023,
  booktitle={Proc. INTERSPEECH 2023},
}

@article{braunschweiler2011automatic,
  title={Automatic sentence selection from speech corpora including diverse speech for improved HMM-TTS synthesis quality},
  author={Braunschweiler, Norbert and Buchholz, Sabine},
  year={2011}
}

@inproceedings{hwang2023pausespeech,
  title={PauseSpeech: Natural speech synthesis via pre-trained language model and pause-based prosody modeling},
  author={Hwang, Ji-Sang and Lee, Sang-Hoon and Lee, Seong-Whan},
  booktitle={Asian Conference on Pattern Recognition},
  pages={415--427},
  year={2023},
  organization={Springer}
}

@article{yang2024frame,
  title={Frame-wise breath detection with self-training: An exploration of enhancing breath naturalness in text-to-speech},
  author={Yang, Dong and Koriyama, Tomoki and Saito, Yuki},
  journal={arXiv preprint arXiv:2402.00288},
  year={2024}
}

@article{Pitt2005TheBC,
  title={The Buckeye corpus of conversational speech: labeling conventions and a test of transcriber reliability},
  author={Mark A. Pitt and Keith Johnson and Elizabeth Hume and Scott F. Kiesling and William D. Raymond},
  journal={Speech Commun.},
  year={2005},
  volume={45},
  pages={89-95},
  url={https://api.semanticscholar.org/CorpusID:3232115}
}

@misc{radford2022robustspeechrecognitionlargescale,
      title={Robust Speech Recognition via Large-Scale Weak Supervision}, 
      author={Alec Radford and Jong Wook Kim and Tao Xu and Greg Brockman and Christine McLeavey and Ilya Sutskever},
      year={2022},
      eprint={2212.04356},
      archivePrefix={arXiv},
      primaryClass={eess.AS},
      url={https://arxiv.org/abs/2212.04356}, 
}
